\documentclass{amsart}
\usepackage{graphicx,amsfonts,amssymb,algorithm}
\theoremstyle{definition}

\theoremstyle{remark}

\numberwithin{equation}{section}

\begin{document}
\title[Pattern recognition issues on anisotropic SPH]{Pattern recognition issues on anisotropic smoothed particle hydrodynamics}%
\author{Eraldo Pereira Marinho}%
\address{Univ Estadual Paulista (UNESP/IGCE), Department of Computing,
 Applied Mathematics and Statistics}%
 \email{emarinho@rc.unesp.br}

\subjclass{Artificial Intelligence on Computational Fluids Dynamics}%

\begin{abstract}
This is a preliminary theoretical discussion on the computational requirements of the state of the art smoothed
particle hydrodynamics (SPH) from the optics of pattern recognition and artificial intelligence. It is pointed out
in the present paper that, when including anisotropy detection to improve resolution on shock layer, SPH is a very
peculiar case of unsupervised machine learning. On the other hand, the free particle nature of SPH opens an
opportunity for artificial intelligence to study particles as agents acting in a collaborative framework in which
the timed outcomes of a fluid simulation forms a large knowledge base, which might be very attractive in
computational astrophysics phenomenological problems like self-propagating star formation.
\end{abstract}
\maketitle
\section{Introduction}

\par
Smoothed particle hydrodynamics (SPH) has been a successful computer simulation paradigm originated in
computational astrophysics since 1977 \cite{Gingold1977,Lucy1977}. Nowadays SPH is used in other areas and has
gained significant improvement in accuracy and stability, not only on simulating compressible shock as also on
performing high resolution incompressible fluid, solids etc, e.g. \cite{Guo2011}.

\par
One essential issue in SPH is anisotropy, which arises naturally on performing adaptive interpolation, e.g.
\cite{marinho2010mecanica}, where the dimensionality reduction to detect critical surfaces, as shock layers might
be included. Anisotropy is an important subject in pattern recognition for feature extraction methods \cite{Duda}.

\par
There are encouraging works on the application of artificial intelligence to reproduce real systems, mainly in the
context of complex systems as in economics and many other social and life sciences, e.g. \cite{VolkerGrimm2005}.
The collective phenomena of star formation, unveiled in the intermittent pattern in the spiral galaxies arms, have
been proposed by means of the stochastic self-propagating star formation model, in the sense that star formation
is contagious, e.g. \cite{Mineikis2010}.

\par
The present work is a brief discussion on some aspects of SPH circumstanced by the concepts of pattern recognition
and artificial intelligence. Several details are omitted to fulfill the limited space with no significative lost
of focus on reviewing the theory in the context of intelligent computing.

\section{SPH database and space representation}

\par
The SPH database comprises $N$ instances, usually thousand hundreds or even millions particles, indexed by a
descriptor table $\mathcal{P}_N$. Each particle is addressed by a unique label, or descriptor $i$, and as much as
possible the particle object is referred as just $i$ or $i$-particle. Any particle attribute, say $A$, is
addressed by sub-indexing the same with the particle label, $A_i$.

\par
Since a specific SPH problem is headed by the mathematical, physical and computational models, the adopted methods
might be included in the database in the form of classes and module libraries, described in a commonly used data
model language as for instance the XML, which may also improve the information interchange between different SPH
simulation bases.

\par
The usual 3D space description in SPH uses hierarchical spatial tessellation, as for instance by means of octrees,
e.g. \cite{marinho2001sph}, which are the 3D version of quadtrees. Other tree-based spatial tessellation schemes
are also adopted. For instance \cite{marinho2010mecanica} proposes an approach to easily adapt earlier versions of
octree-based SPH codes to covariance-based octree tessellation to improve anisotropic kernel computations, e.g.
\cite{2006TWB.51.513-525}.

\section{Starring pattern recognition and AI in SPH}


\par
The $k$-nearest neighbor is a mathematical relation $\mathcal{N}_k\subseteq\mathcal{P}_N\times\mathcal{P}_N$,
which associates $i\in\mathcal{P}_N$ with a subset $\mathcal{N}_k(i)=\{i_1,\ldots,i_k\}\subseteq\mathcal{P}_N$, so
that $j\in\mathcal{N}_k(i)$ if, and only if, the adopted distance $d(\vec x_i,\vec x_j)$ from $i$ to $j$ obeys the
inequality $d(\vec x_i,\vec x_j)\leq\max\{d(\vec x_i,\vec x_l)|\,l\in\mathcal{N}_k(i)\}$. The KNN algorithm is the
method by which the relation $\mathcal{N}_k$ is populated by ordered pairs $(i,j)$ in
$\subseteq\mathcal{P}_N\times\mathcal{P}_N$, given the particle-descriptor table $\mathcal{P}_N$.

\par
The $\mathcal{N}_k$ relation is asymmetric and reflexive. The later comes from the fact that each point is the
nearest neighbor of itself -- this is called improper neighbor. The former comes from the fact that if $a$ is the
proper nearest neighbor (not a reflection) of $b$, not necessarily $b$ is the nearest neighbor of $a$. For
example, $a$ could be closer to a third point $c$ than $b$, which is too faraway from any other point but $a$.

\par
The KNN asymmetry reflects imperfections on writing simpler forms of the SPH conservation equations, which require
particle commutation symmetry. To workaround the asymmetry issue is necessary to introduce the {\em symmetric
closure} of the KNN relation, which is known as effective neighbors
\begin{equation}\label{eq:eff-neighb}
\mathcal{E}_k=\mathcal{N}_k\cup\{(i,j)\in\mathcal{P}_N\times\mathcal{P}_N\;|\;(j,i)\in\mathcal{N}_k\}.
\end{equation}

\par
Of course, the KNN algorithm requires a predesign metric in the 3D space. If the metric is invariant under
rotation, the KNN relation is isotropic. On the other hand, the metric relation is said anisotropic. For instance
the Mahalanobis metric, as adopted in the KNN algorithm proposed by \cite{marinho2010mecanica}, is anisotropic and
is used to reveal biased structures like the arms in spiral galaxy images.

\par
The Mahalanobis distance $\xi_{ij}$ is defined in terms of the covariance tensor $\mathbf{\Sigma}$:
\begin{equation}\label{eq:mahalanobis-distance}
\xi_{ij}=(\vec x_i-\vec x_j)\mathbf{\Sigma}^{-1}(\vec x_i-\vec x_j)^{\mathrm{T}},
\end{equation}
where $(\vec x_i-\vec x_j)^{\mathrm{T}}$ is the transpose of the matrix representation of the relative position
vector $(\vec x_i-\vec x_j)$.

\par
Of course, equation~(\ref{eq:mahalanobis-distance}) is not the only way of defining anisotropic distance in SPH.
For instance, the positive-definite stress tensor $\mathbf{T}$ might be eventually used to define the
non-normalized anisotropic distance $\xi_{ij}$:
\begin{equation}\label{eq:stress-distance}
\xi_{ij}=(\vec x_i-\vec x_j)\mathbf{T}^{-1}(\vec x_i-\vec x_j)^{\mathrm{T}}.
\end{equation}

\par
According to equations~(\ref{eq:mahalanobis-distance}) or (\ref{eq:stress-distance}), the outermost boundary for
the $k$-nearest neighbors of the $i$-particle is an ellipsoid centered in the query position $\vec x_i$, whose
principal axes are set by the respective tensor eigenvectors \cite{marinho2010mecanica}.


\par
A cognitive interpretation for the well-known SPH interpolation formula can be illustrated as follows: given an
$i$-labeled particle, say $i$-particle, one may suppose this particle has to make an estimation, $\tilde{A}_i$, of
a local fluid quantity, $A_i$, after hearing votes, e.g. \cite{Duda}, from its effective neighbors what impression
they get regarding the same quantity.

\par
A democratic decision is made if the $i$-particle weights the individual suggestions from its informants, giving
more importance to the closest ones. The importance, or weight, comes from a compact-support smoothing kernel,
which drops to zero outside the influence zone defined by the effective neighbors and grows up as gets closer to
$i$, reaching its maximum for $i$ itself.

\par
Each $i$-particle, $i=1,\ldots,N$, has its own effective neighbors, $\mathcal{E}_k(i)$. The $i$-particle asks each
$j$-particle in $\mathcal{E}_k(i)$ for suggestions, which answers accordingly to the predefined protocol,
$A_j{m_j}/{\rho_j}$, whose reliability is expressed by a weight, or smoothing kernel $W_{ij}$.

\par
The $i$-particle gets a conclusive perception $\tilde{A}_i$ from its locality by adding together all of the
weighted votes, $W_{ij}A_j{m_j}/{\rho_j}$, received from its $k$-nearest neighbors:
\begin{equation}\label{eq:p-interpol}
\tilde{A}_i=\sum_{\forall j\in\,\mathcal{E}_k(i)}W_{ij}A_j\frac{m_j}{\rho_j},
\end{equation}
where $W_{ij}=W(\vec x_i-\vec x_j)$ is the smoothing kernel, whose analytical profile might be an issue regarding
accuracy and stability on SPH simulations, but this particular subject will not be discussed here.

\par
Similar election procedure applies on estimating the interpolated gradient, $\vec\nabla_i{A}_i$, yielding
\begin{equation}\label{eq:grad-interpol}
\vec\nabla_i{A}_i=\sum_{\forall j\in\,\mathcal{E}_k(i)}\vec\nabla_i{W}_{ij}A_j\frac{m_j}{\rho_j},
\end{equation}
where $\vec\nabla_i{W}_{ij}=\vec\nabla_i{W}(\vec x_i-\vec x_j)$ is the smoothing-kernel gradient.

\par
If the kernel is symmetric, one finds from the effective neighbors symmetry that
$i\in\mathcal{E}_k(j)\Leftrightarrow j\in\mathcal{E}_k(i)$, and also finds $W_{ij}=W_{ji}\neq0$, and
$\vec\nabla_i{W}_{ij}=-\vec\nabla_j{W}_{ji}$ if and only if $(i,j)\in\mathcal{E}_k$.

\par
Densities are required to perform SPH interpolations, as in equations~(\ref{eq:p-interpol}) and
(\ref{eq:grad-interpol}), and they are estimated from equation~(\ref{eq:p-interpol}) itself by means of a
self-consistent replacement $A_j\rightarrow\rho_j$, yielding
\begin{equation}\label{eq:dens-interpol}
\tilde{\rho}_i=\sum_{\forall j\in\,\mathcal{E}_k(i)}W_{ij}{m_j}=\rho_i.
\end{equation}


\par
The SPH fluid equations of motion are derived from the actual fluid equations, and they must be solved by means of
some integration scheme regarding accuracy and stability. The timed outcomes from the integration scheme express
discrete states of the particle description. Depending on the time-integration method, each particle knows a brief
history of its previous states.

\par
The way as the SPH equations are presented usually requires rearrangement to attend to subsidiary information
concerning physics, chemistry etc. For instance, in most astrophysical problems, the SPH momentum conservation
equation can be written as
\begin{equation}\label{eq:sph-momentum}
\frac{\mathrm{d}\vec v_i}{\mathrm{d}t}=-\sum_{\forall j\in\,{\mathcal{E}_k(i)}}\vec{\nabla}_iW_{ij}\Pi_{ij} +
\vec{F}_i,
\end{equation}
where the $\Pi_{ij}$-factor carries the pressure coefficients, which might even include anisotropic pressure as
the elastic stress tensor and the Maxwell stress tensor, e.g. \cite{marinho2001sph}. The $\vec{F}_i$-vector term
is a non-hydrodynamic acceleration as for instance the gravity field $\vec{F}_i=-\vec g_i$ on $i$-particle.


\par
Time-integration scheme plays the role of particle actuators modifying their local environment, in response to the
information received from their effective neighbors. Every particle contributes to a global knowledge, which might
attend to a subsidiary simulation, as for example the qualitative results of self-propagating star formation, e.g.
\cite{Mineikis2010}, the SPH-data history constitutes a knowledge-base \cite{Russell2009} or even a more
pretentious context as in live tissue simulations \cite{Muller:2004:IBS:1011146.1011149}.

\par
Each SPH particle recognizes its surroundings by means of its effective neighbors using pattern detection
techniques to identify the neighborhood morphology and consequent critical surfaces. However, particles obey a set
of transition rules, according to the physics model, to decide what action they have to do against their local
environment.

\par
From the theory of intelligent agents, SPH particles might be classified as {\em simple reflex agents}
\cite{Russell2009}, acting as environment modifiers in function of what they percept in their surrounds through
their effective neighbors. The particles act under the local physical conditions in response to the input they
receive from their effective neighbors, ignoring the long term history of all their actions and percepts.
Regarding the adopted time integration scheme embedded as actuators, only the knowledge of a recent past is
required.

\section{Conclusion}

\par
More than a numerical simulation technique, SPH is a very complex system that can be studied not only under
applied mathematics techniques but also under the light of intelligent computing, where particles are individuals
cooperatively working in behalf of a collective objective of mimicking the fluid behavior.

\par
The SPH spirit resides in computationally reproducing the continuous fluid flow using free particles. A fluid
particle moves like a marker, accordingly to the lagrangian equations of motion. Each particle is a data structure
storing the specific fluid properties as density, pressure, position, velocity etc. Any particle knows its
surroundings through its $k$-nearest neighbors (KNN), which play the role of sensors, or informants. The
information mechanism is known as KNN-based kernel interpolation, which might be interpreted as a weighted voting,
from the machine learning viewpoint.

\bibliographystyle{amsplain}

\end{document}